\pdfoutput=1
\documentclass[letterpaper]{article}
\usepackage{aaai21}
\usepackage{times}
\usepackage{helvet}
\usepackage{courier}
\usepackage{graphicx}

\usepackage{natbib}
\usepackage{caption}
\usepackage{xcolor}
\usepackage{lineno}
\usepackage{subfigure}
\usepackage{amsmath}
\usepackage{multirow}
\usepackage{tabularx}
\usepackage{rotating}
\usepackage{booktabs}
\usepackage{makecell}
\usepackage{verbatim}
\usepackage{amssymb}
\usepackage{xcolor}
\usepackage{comment}
\usepackage{cleveref}
\newcolumntype{P}[1]{>{\centering\arraybackslash}p{#1}}
\def\ie{\emph{i.e.}}
\def\eg{\emph{e.g.}}
\def\etc{\emph{etc}}

\def\ourmodel{\emph{RD3D}}

\begin{document}
%
\title{RGB-D Salient Object Detection via 3D Convolutional Neural Networks}
\author{Qian Chen\textsuperscript{\rm 1}, Ze Liu\textsuperscript{\rm 1}, Yi Zhang\textsuperscript{\rm 2}, Keren Fu\textsuperscript{\rm 3,4}\thanks{Corresponding author: Keren Fu \emph{(fkrsuper@scu.edu.cn)}}, Qijun Zhao\textsuperscript{\rm 3,4}, Hongwei Du\textsuperscript{\rm 1}\\
}
\affiliations{
    \textsuperscript{\rm 1}School of Information Science and Technology, University of Science and Technology of China\\ 
    \textsuperscript{\rm 2}Institut National des Sciences Appliqu\'ees de Rennes\\
    \textsuperscript{\rm 3}College of Computer Science, Sichuan University\\
    \textsuperscript{\rm 4}National Key Laboratory of Fundamental Science on Synthetic Vision, Sichuan University\\
}
\maketitle
\begin{abstract}
\begin{quote}

RGB-D salient object detection (SOD) recently has attracted increasing research interest and many deep learning methods based on encoder-decoder architectures have emerged. However, most existing RGB-D SOD models conduct feature fusion either in the single encoder or the decoder stage, which hardly guarantees sufficient cross-modal fusion ability.
In this paper, we make the first attempt in addressing RGB-D SOD through 3D convolutional neural networks. The proposed model, named \ourmodel, aims at pre-fusion in the encoder stage and in-depth fusion in the decoder stage to effectively promote the full integration of RGB and depth streams.
Specifically, \ourmodel~first conducts pre-fusion across RGB and depth modalities through an inflated 3D encoder, and later provides in-depth feature fusion by designing a 3D decoder equipped with rich back-projection paths (RBPP) for leveraging the extensive aggregation ability of 3D convolutions. With such a progressive fusion strategy involving both the encoder and decoder, effective and thorough interaction between the two modalities can be exploited and boost the detection accuracy. Extensive experiments on six widely used benchmark datasets demonstrate that \ourmodel~performs favorably against 14 state-of-the-art RGB-D SOD approaches in terms of four key evaluation metrics. Our code will be made publicly available: https://github.com/PPOLYpubki/RD3D.
\end{quote}
\end{abstract}

\section{Introduction}
Salient object detection (SOD) aims to imitate the human visual system on detecting objects or areas that attract human attention~\cite{jiang2020light,fan2018salient,zhao2019egnet}. SOD has a wide range of applications in many tasks, such as object segmentation and recognition~\cite{han2005unsupervised,li2011saliency}, video detection~\cite{li2019motion,fan2019shifting}, content-related image and video compression~\cite{itti2004automatic,guo2009novel} as well as tracking~\cite{zhang2020non}. Although SOD has been advanced notably by deep learning techniques~\cite{wang2019salient}, single-modal SOD still faces many problems, such as weak appearance differences in the foreground and background regions, complex foreground and background, \etc.

\begin{figure}
	\centering
	\includegraphics[width=1.0\linewidth]{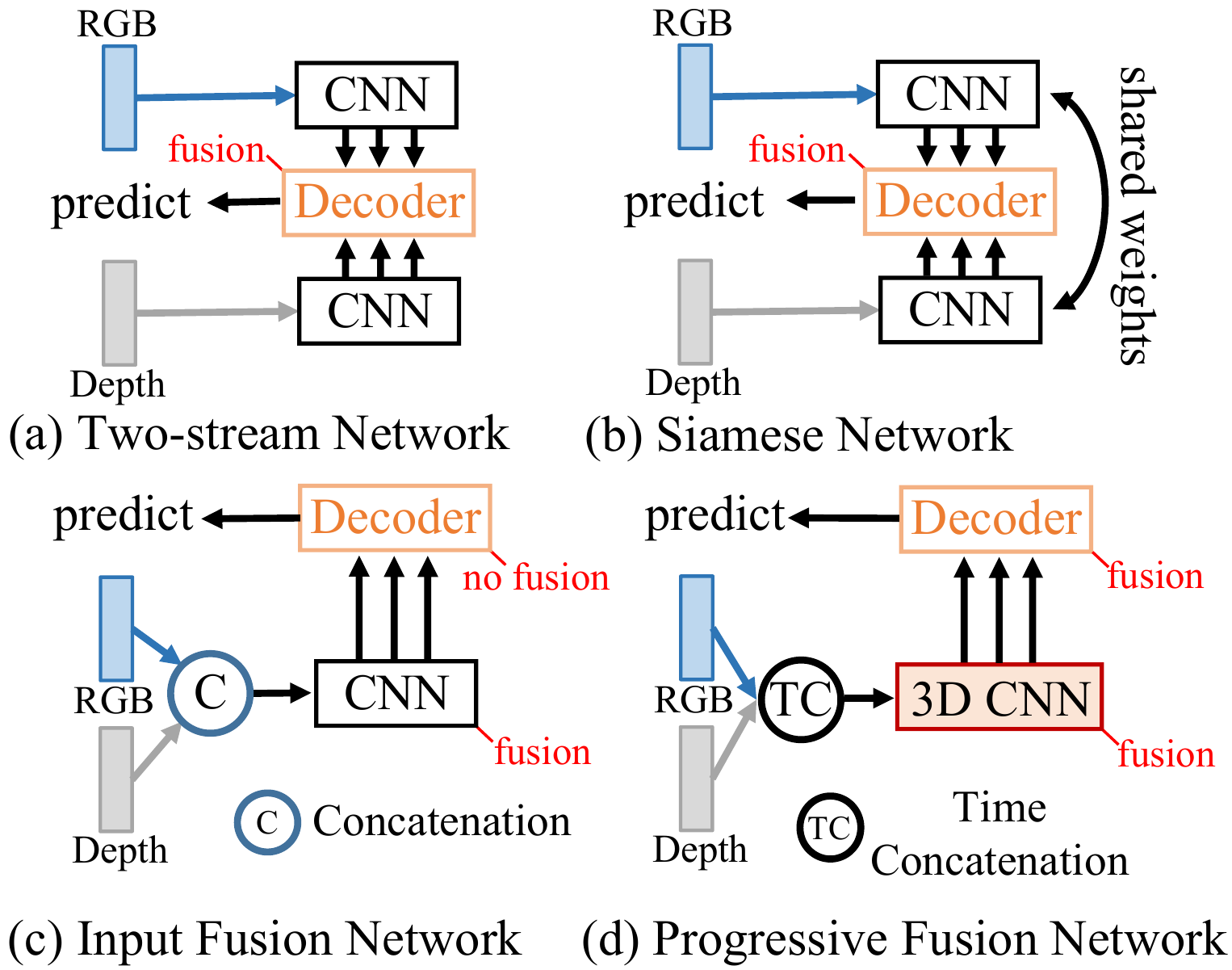}\vspace{-0.3cm}
	\caption{\small Categorization of existing models. Note that (a)-(c) conduct feature fusion either in the encoder or the decoder stage, while our model (d) adopts progressive fusion involving both the encoder and decoder stages.}\vspace{-0.8cm}
	\label{related}
\end{figure}

In recent years, an increasing number of RGB-D SOD models have emerged to address these challenges of single-modal SOD for more accurate detection performance \cite{zhang2020bilateral}. Although encouraging results have been obtained, we notice that existing models conduct feature fusion either in the single encoder or the decoder stage, which may hardly guarantee sufficient cross-modal fusion ability. As shown in Fig. \ref{related}, these models can be divided into three categories according to how they extract and fuse cross-modal features.
In the first category (Fig. \ref{related} (a)), the models~\cite{fan2020bbs,pang2020hierarchical,liu2020learning,piao2020a2dele,zhang2020select} extract features from RGB and depth maps independently, and conduct feature maps fusion of the two modalities in the decoder. To achieve effective cross-modal fusion, the authors tend to elaborately design complex or special modules for simultaneous fusion and decoding.
The second category \cite{fu2020jl,li2020icnet} (Fig. \ref{related} (b)) uses a Siamese network as an encoder to extract features from RGB and depth. Although the encoder network is shared across different modalities, however, it is still dedicated to feature extraction similar to Fig. \ref{related} (a) and no fusion behavior is conducted in the encoder.
The third category of models \cite{zhao2020single,fan2020rethinking,song2017depth,liu2019salient} (Fig. \ref{related} (c)) adopt the ``input fusion'' strategy, which concatenates RGB and depth across channel dimension before feeding them to the encoder. In this case, the main role of fusion is played by the encoder since all the ingredients fed to the decoder are already-fused features, making the decoder infeasible to conduct explicit cross-modal fusion.

Considering that feature extraction and fusion is crucial in such an encoder-decoder architecture for the RGB-D SOD task, the aforementioned models have not fully investigated the feature aggregation potentials in both the encoder and decoder. Inspired by the success of 3D convolutional neural networks (CNNs) in aggregating extensive feature information for space-time processing (\eg, video recognition~\cite{feichtenhofer2020x3d}, action localization~\cite{gu2018ava}) where 3D CNNs often serve as encoders, we propose to treat the depth modality as another ``time state'' of the RGB one and aggregate information of the two modalities through 3D CNNs. To the best of knowledge, our work is \emph{the first attempt that addresses RGB-D SOD through 3D CNNs}, attributed to which RGB and depth information can be mutually enhanced meanwhile making explicit fusion in the decoder possible. Another advantage is that due to the inner fusion behavior of 3D convolutions, dedicated or sophisticated modules for cross-modal fusion are \emph{no longer required}.
The proposed novel model, named \ourmodel~(short for \textbf{R}GB-\textbf{D} \textbf{3D} CNN detector for SOD), first conducts pre-fusion across RGB and depth modalities through an inflated 3D encoder. Then, the obtained pre-fused RGB and depth features are fed to a 3D decoder for further in-depth fusion. The 3D decoder incorporates rich back-projection paths (RBPP) in order to better leverage the extensive aggregation ability of 3D convolutions. Therefore, both the encoder and decoder of \ourmodel~are 3D CNNs-based and they both involve cross-modal fusion in a progressive manner (Fig. \ref{related} (d)). Our work has three main contributions:
\begin{itemize}
\item We exploit the idea of pre-fusion in the encoder stage and show how it is beneficial to the final performance. We propose to tackle this by 3D CNNs, which can fuse the cross-modal features effectively without requiring dedicated or sophisticated modules.

\item We design a 3D decoder that incorporates rich back-projection paths (RBPP) in order to better leverage the extensive aggregation ability of 3D convolutions. Such a 3D decoder makes the proposed \ourmodel~a fully 3D CNNs-based model and also the first 3D CNNs-based model for the RGB-D SOD task.

\item We show that \ourmodel, which is the first 3D CNNs-based model for RGB-D SOD, surpasses 14 state-of-the-art (SOTA) methods by a notable margin on the six widely used benchmark datasets.
\end{itemize}

\begin{figure*}[!htb]
	\centering
	\includegraphics[width=1\linewidth]{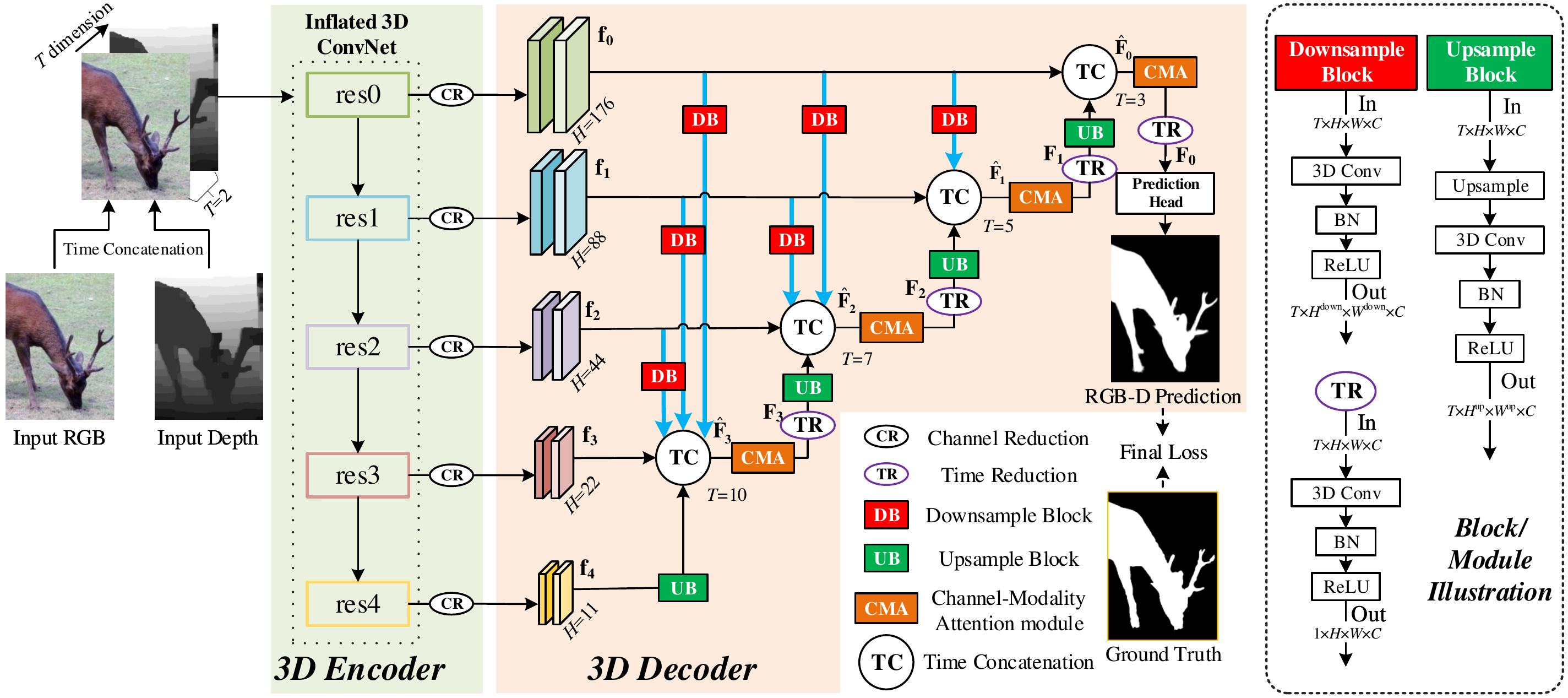}\vspace{-0.4cm}
	\caption{\small Block diagram of the proposed \ourmodel~scheme for RGB-D SOD. $H$ denotes the spatial resolution of output feature maps at each level, whereas $T$ denotes the temporal dimension. Definitions of $\mathbf{f}_{i}$, $\mathbf{\hat F}_{i}$, and $\mathbf{F}_{i}$ can be found in Eq. (\ref{equ2}) and Eq. (\ref{equ3}).}\vspace{-0.4cm}
	\label{label1}
\end{figure*}

\section{Related Work}

\noindent \textbf{Traditional RGB-D Models.} Previous algorithms/models in the RGB-D SOD field focus on using hand-crafted features, such as contrast~\cite{cong2016saliency,cheng2014depth,desingh2013depth}, center or boundary prior~\cite{cheng2014depth,cong2019going}, center-surround difference~\cite{ju2014depth,guo2016salient}, \etc. These methods heavily rely on hand-crafted features and lead to poor performance in complex scenes.

\noindent \textbf{Deep-based RGB-D Models.} Existing models can be divided into three classes according to the stage of fusion: early-fusion~\cite{peng2014rgbd,song2017depth}, middle-fusion~\cite{feng2016local,fu2020jl,fu2020siamese,zhang2020uc,piao2019depth} and late-fusion~\cite{fan2014salient}. By contrast, as shown in Fig~\ref{related}, this paper elaborately divides current methods into four categories according to how they extract and fuse cross-modal features. Among the first category named the two-stream network (Fig. \ref{related} (a)), \cite{han2017cnns} utilized a CNN network to extract information from the two modalities in the backbone stage, and then fused such deep representations from multi-views via a fully connected layer. \cite{piao2019depth} proposed a novel depth-induced multi-scale recurrent attention network, which extracted features respectively from RGB and depth maps and then input them to depth refinement blocks for integration. ~\cite{chen2020ef} utilized separate CNNs to extract features from RGB and depth modalities. The resulting hint map is then utilized to enhance the depth map, which suppresses the noise and sharpens the object boundary. The second category is the Siamese network (Fig. \ref{related} (b)). Fu \textit{et al.} \cite{fu2020jl} and Li \textit{et al.} \cite{li2020icnet} first adopted a Siamese network with shared weights for the RGB/depth stream during independent feature extraction. The third category is called the input fusion network (Fig. \ref{related} (c)). \cite{huang2018rgbd} and \cite{liu2019salient} concatenated RGB and depth maps to formulate a four-channel input, which was fed to a single-stream CNN. DANet proposed by Zhao~\textit{et al.}~\cite{zhao2020single} fused bi-modal information in the input stage, and meanwhile depth maps played a guidance role in the decoder stage. In general, the above representative works do their utmost to explore: 1) effective utilization of depth information, and 2) comprehensive fusion of RGB and depth cues. Unfortunately, they have the limitation that feature aggregation potentials in both the encoder and decoder are not fully leveraged. Complete survey of models in this field can be found in \cite{zhou2020rgbd}. Different from existing models, we propose the 3D CNNs-based progressive fusion scheme (Fig. \ref{related} (d)) towards a new perspective of multi-modal feature extraction and fusion.

\noindent \textbf{3D CNNs.} 3D CNNs are influential in many fields, such as video processing~\cite{ji20123d,tran2015learning}, medical image processing~\cite{balakrishnan2019voxelmorph} and point cloud processing~\cite{zhou2018voxelnet}. Balakrishnan~\textit{et al.}~\cite{balakrishnan2019voxelmorph} applied 3D convolutions to extract features from image volumes in the encoder stage and utilized a 3D CNN-based decoder to transform features on finer spatial scales, enabling precise anatomical alignment. By using 3D convolutions, Zhou~\textit{et al.}~\cite{zhou2018voxelnet} extracted features from 3D voxels for point cloud-based 3D object detection. The above methods use 3D convolutions to handle data resided in the 3D space, but 3D convolutions can also process data in multi-domain. Ji~\textit{et al.}~\cite{ji20123d} applied them to extract features in spatial and temporal domains from video data to capture motion information. To the best of our knowledge, we are the first to investigate 3D convolutions for RGB-D saliency detection.

\section{Methodology}
\subsection{Big Picture}
The overall architecture of the proposed \ourmodel~is shown in Fig. \ref{label1}. It follows the typical encoder-decoder architecture and is composed of a 3D encoder and a 3D decoder. The 3D encoder is basically a ResNet/VGG-like backbone which is extended by 3D convolutions. It aims at cross-modal feature pre-fusion while its outputs are modality-aware multi-level features. On the other hand, the 3D decoder decodes features by 3D convolutions. It follows the typical UNet-like top-down fashion but incorporates rich back-projection paths (RBPP, the blue line arrows in Fig. \ref{label1}) as well as channel-modality attention modules (CMA, the orange modules in Fig. \ref{label1}). After the final decoding by 3D convolutions, the decoder outputs a prediction map highlighting salient object(s). Noting that attributed to the extensive aggregation ability of 3D convolutions, no any explicit cross-modal fusion modules are used in Fig. \ref{label1}.

\subsection{3D Encoder}
As shown in Fig~\ref{label1}, given an RGB image and a single-channel depth map, we first normalize the depth map into intervals $[0,255]$ and then replicate it into three channels. Hereafter, we follow~\cite{wang2018non} and denote the dimension of a tensor as $T\times H\times W\times C$, where ``$T$'' refers to the temporal dimension and ``$H$'', ``$W$'', ``$C$'' mean the height, width, and channels, respectively. We stack the RGB image ($H\times W \times C$) and the corresponding depth map ($H\times W \times C$) to form a 4D tensor as the input of our 3D encoder, where $T=2$ and $C=3$.
\begin{figure}
	\centering
	\includegraphics[width=.8\linewidth]{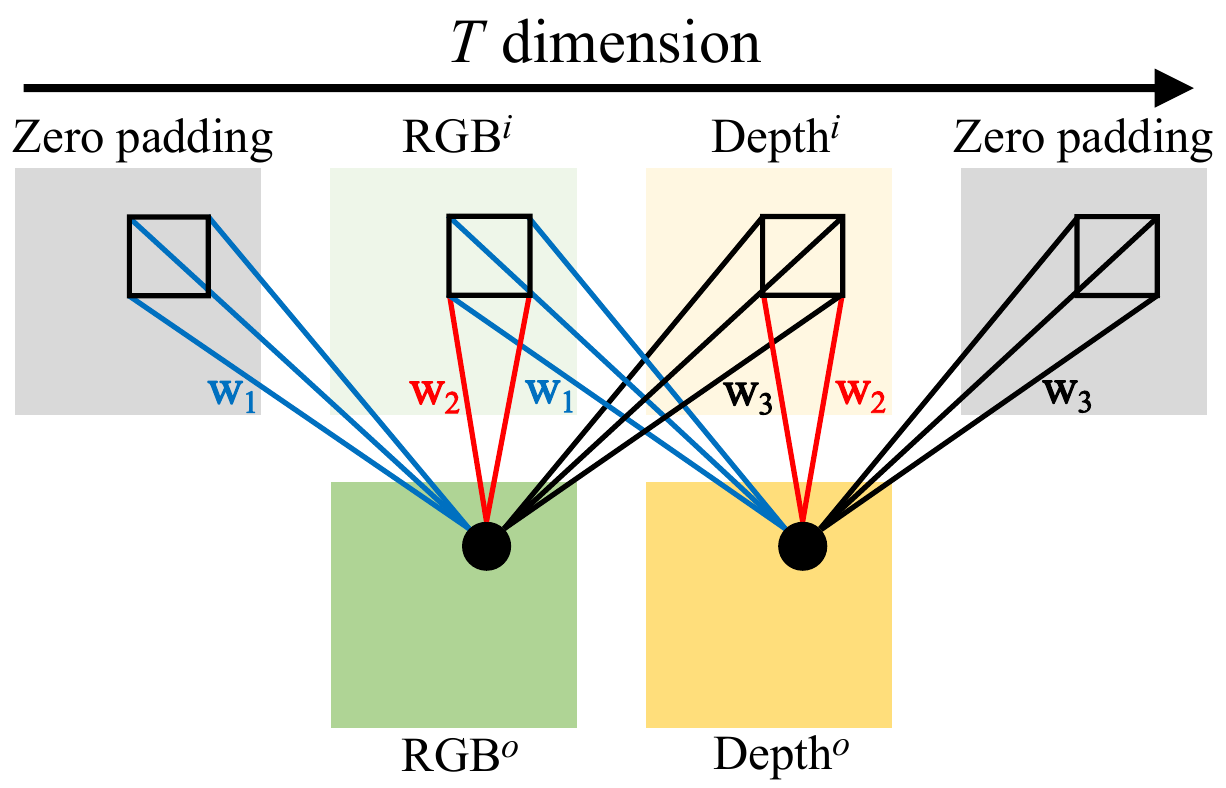}\vspace{-0.4cm}
	\caption{\small Visualization of the 3D convolution in the ``$T$'' dimension, where the corresponding kernel size is 3. The superscript ``$i/o$'' means input/output features of the 3D convolution.}\vspace{-0.5cm}
	\label{3dconv}
\end{figure}
We adopt an inflated 3D ResNet~\cite{carreira2017quo} as our encoder, which replaces all 2D convolutions in the conventional ResNet~\cite{ResNET} with 3D convolutions, and the kernel sizes for the ``$T$'' dimension are set as 3 for all the $3 \times 3$ 3D convolutions, with padding, stride, output dimension being 1, 1 and 2, respectively. Computation in the ``$T$'' dimension of a 3D convolutional layer thus can be visualized in Fig. \ref{3dconv} and is equivalent to the formulations below:

\vspace{-0.2cm}
\begin{equation}\label{equ1}
\begin{split}
    \mathbf{R}^o=\mathbf{w_2} \ast \mathbf{R}^i + \mathbf{w_3} \ast \mathbf{D}^i,\\
    \mathbf{D}^o=\mathbf{w_1} \ast \mathbf{R}^i + \mathbf{w_2} \ast \mathbf{D}^i,
\end{split}
\end{equation}
where $\mathbf{w_1}$, $\mathbf{w_2}$ and $\mathbf{w_3}$ represent the three temporal weight slices of the 3D kernel. $\mathbf{R}^i$ and $\mathbf{D}^i$ denote the input RGB and depth feature slices, respectively, whereas $\mathbf{R}^o$ and $\mathbf{D}^o$ denote the output ones. ``$\ast$'' means the 2D convolution operation. One can see that the inner fusion property of 3D convolutions helps fuse the RGB and depth information, where RGB and depth cues are mutually enhanced by each other when passing through a 3D convolutional layer. Therefore, progressive fusion can be achieved when using successive 3D convolutions.

\begin{figure}
	\centering
	\includegraphics[width=1.0\linewidth]{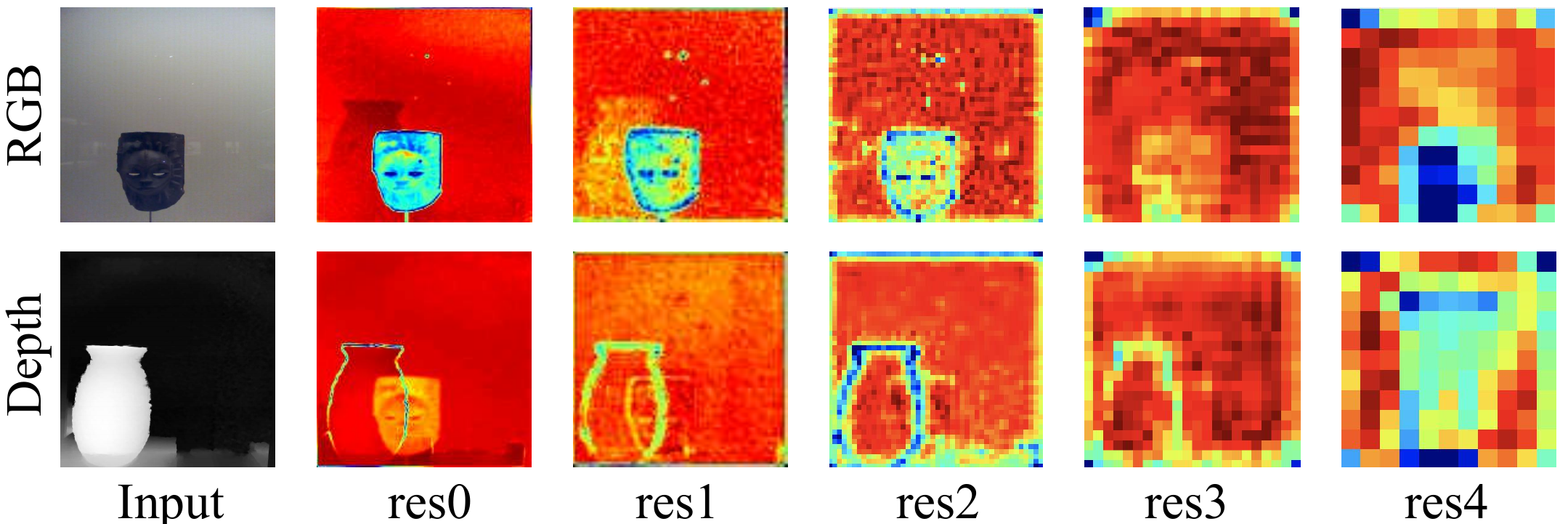}\vspace{-0.3cm}
	\caption{\small Modality-aware hierarchical features in each temporal slice. To make the impact of pre-fusion more visually obvious, we intentionally feed to our encoder RGB and depth images that are not matched. As a result, explicit fusion behavior can be observed.}\vspace{-0.3cm}
	\label{label3}
\end{figure}

Also note that the output number in the ``$T$'' dimension of our encoder is fixed as 2 under the particular consideration that there are only two modalities in our problem, namely RGB and depth. Although there exist other temporal designs of 3D kernels, Eq. (\ref{equ1}) is adequate to reflect our idea of using 3D CNNs. Specifically, in Eq. (\ref{equ1}) RGB and depth cues are preserved by shared weights $\mathbf{w_2}$, and meanwhile each one is enhanced by the other by learnable weights $\mathbf{w_1}$/$\mathbf{w_3}$. This achieves certain modality-aware individuality as well as cross-modal fusion, leading to the term ``\emph{pre-fusion}''. Fig. \ref{label3} visualizes either temporal slice of feature maps at different levels, from which it is observed that information between the two modalities are cleverly integrated, but they are not the same. Finally, as shown in Fig~\ref{label1}, the yielded modality-aware multi-level features whose temporal dimensions equal to 2 are fed to channel reduction (CR) modules to reduce their channels to a fixed smaller number (while the other dimensions are unchanged), \ie~32 in practice, for the subsequent decoding. This is to reduce computation load as well as memory usage.

Inspired by~\cite{carreira2017quo,feichtenhofer2016spatiotemporal,girdhar2018detect}, we propose to initialize our 3D encoder in a centralized strategy using ImageNet pre-trained weights of the 2D ResNet, namely using such 2D weights to initialize the central slice $\mathbf{w_2}$ of a 3D kernel while setting other slices to 0, \ie, $\mathbf{w_1}=\mathbf{w_3}=0$. This strategy is equivalent to using a shared 2D ResNet to process RGB and depth at the beginning, which exactly coincides with the recent idea of using Siamese network for RGB-D SOD \cite{fu2020jl}.


\subsection{3D Decoder with Rich Back-Projection Paths}
As shown in the decoder part of Fig. \ref{label1}, the channel-reduced features at each spatial resolution is aggregated with those at other resolutions in a hierarchical way. Inspired by but different from the widely employed UNet-like top-down fashion, which only considers \emph{upsampling} low-resolution features to incorporate with high-resolution ones for refinement, we propose to combine additional  \emph{downsampling} flows from high-resolution features to low-resolution ones, denoted by the blue line arrows in Fig. \ref{label1}, to leverage the extensive aggregation ability of 3D convolutions. Such downsampling flows transport rich feature information from the higher resolutions to the lower resolutions, enriching high-level feature representation. Note that besides the classical UNet architecture, this also contrasts to the existing technique~\cite{hou2017deeply} whose short connections transport information from low-level to high-level, since we transport in the opposite direction. We call our this method Rich Back-Projection Paths (RBPP). Another important reason of using RBPP is that, 3D convolutions will be more memory- and computation-efficient when used in such a decoder than in RBPP's counterparts that transport features in the opposite direction, like in~\cite{hou2017deeply}.

To be more specific in Fig. \ref{label1}, for the $i$th level, we use a series of downsampling blocks to back-project features from all higher resolutions and meanwhile use an upsampling block to upsample the nearby aggregated feature outputs. The downsampling block is composed of a $1\times3\times3$ 3D convolutional layer, a BN layer, and a ReLU layer. In contrast, the upsampling block is composed of a bilinear upsampling layer and a $1\times3\times3$ 3D convolutional layer, followed by a BN layer and a ReLU layer. Note that both the downsampling and upsampling blocks will keep the temporal dimension number unchanged. Below, we denote the two blocks as $DB(\cdot)$ and $UB(\cdot)$, respectively. The feature computation at the $i$th level ($i\in\{0,1,2,3\}$) is formulated as:

\vspace{-0.2cm}
\begin{align}
   & \mathbf{\hat F}_{i} =  TConcat(\mathbf{f}_{i}, DB(\mathbf{f}_{i-1})...DB(\mathbf{f}_{0}), UB(\mathbf{F}_{i+1}))\label{equ2}\\
   & \mathbf{F}_{i}=TR(CMA(\mathbf {\hat F}_{i})),
	\label{equ3}
\end{align}

\noindent where $TConcat(\cdot)$ means time concatenation (\ie, concatenating in the temporal axis), $\mathbf{f}_{i}$ means the $i$th-level reduced feature tensor after the CR module in the encoder, $\mathbf{F}_{i+1}$ is the nearby feature outputs computed at the $(i+1)$th level, $TR(\cdot)$ denotes a temporal reduction operation which reduces the temporal dimension number to 1 as shown in Fig. \ref{label1}, and $CMA(\cdot)$ denotes the Channel-Modality Attention module introduced below. $\mathbf{\hat F}_{i}$ denotes the intermediate features whereas $\mathbf{F}_{i}$ is the final feature outputs at the $i$th level. Note that we set $\mathbf{F}_{4}=\mathbf{f_{4}}$, and the kernel sizes of the 3D convolutions in $TR(\cdot)$ vary from $10\times 1 \times 1$ (when $i=3$) to $3\times 1 \times 1$ (when $i=0$). After $\mathbf{F}_{0}$ is obtained, a prediction head consisting of a ($1\times 1\times 1,1$) convolutional layer and a Sigmoid layer is used to get the final prediction map.

\begin{figure}
	\centering
	\includegraphics[width=1.0\linewidth]{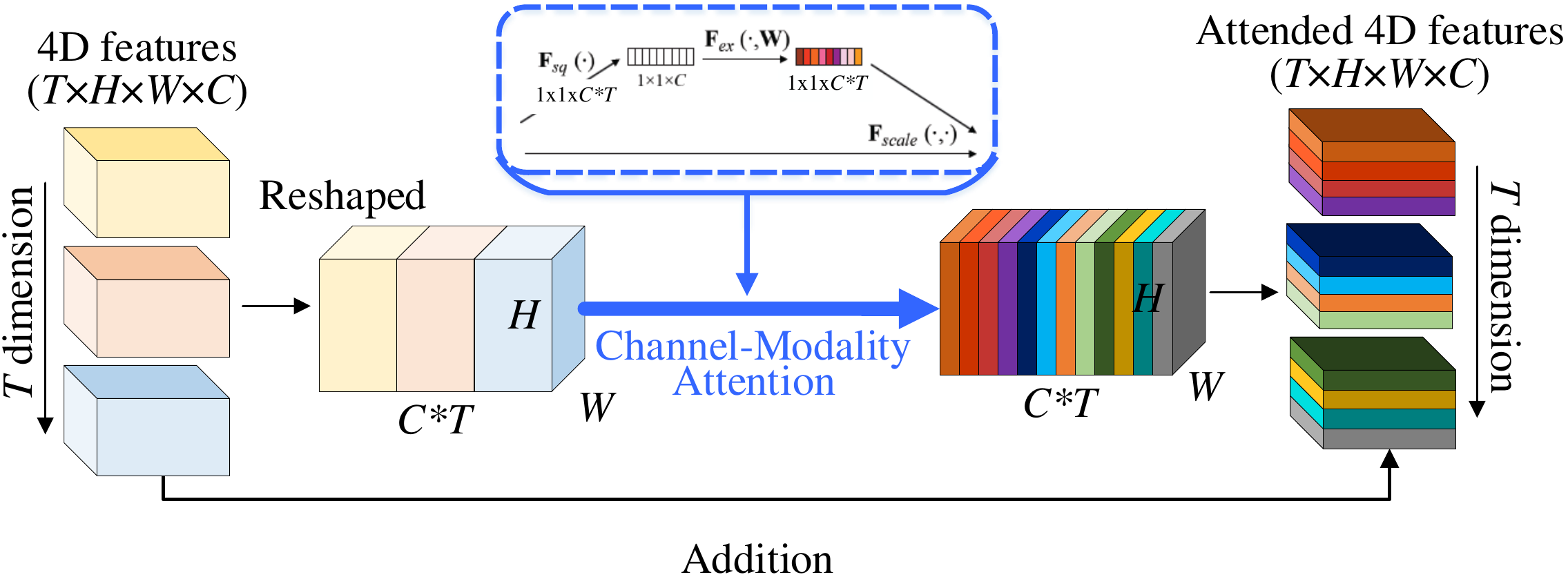}\vspace{-0.3cm}
	\caption{\small Proposed 3D channel-modality attention module that attends on both channel and temporal dimensions.}\vspace{-0.3cm}
	\label{label2}
\end{figure}
\begin{table*}[t!]
	\centering
	\caption{\small Quantitative SOD results in terms of S-measure ($S_\alpha$), maximum F-measure ($F_\beta^{\rm max}$), maximum E-measure ($E_\phi^{\rm max}$) and mean absolute error ($\mathcal{M}$). Six widely used benchmark datasets are employed in the evaluation.
	$\uparrow$/$\downarrow$ denotes that a larger/smaller value is better. The best results are highlighted in \textbf{bold}.}\vspace{-0.5cm}
	\label{table:QuantitativeResults}
	\vspace{8pt}
    \scriptsize
	\renewcommand{\arraystretch}{0.7}
	\renewcommand{\tabcolsep}{1.07mm}
	\begin{tabular}{lr|cccccccccccccc||c}
		\hline\toprule
		& \multirow{2}{*}{Metric}   & AFNet & CTMF & PCF & MMCI & CPFP & D3Net & DMRA  &SSF &A2dele  & JLDCF &UCNet &CoNet &cmMS  &DANet  & \textbf{RD3D} \\
		&   &\scriptsize Access19 &\scriptsize TOC18   &\scriptsize CVPR18 &\scriptsize PR19 &\scriptsize CVPR19 &\scriptsize TNNLS20 &\scriptsize ICCV19  &\scriptsize CVPR20 &\scriptsize CVPR20  &\scriptsize CVPR20 &\scriptsize CVPR20  &\scriptsize ECCV20  &\scriptsize ECCV20 &\scriptsize ECCV20 & \textbf{Ours} \\
		\hline
		\multirow{4}{*}{\begin{sideways}\textit{NJU2K}\end{sideways}}
		& $S_{\alpha}\uparrow$  & 0.772 & 0.849 & 0.877 & 0.858  & 0.879 & 0.893 & 0.886  & 0.899 &0.869 & 0.903 &0.897 &0.895 &0.900  &0.899  & \bf{0.916}\\
		& $F_{\beta}^{\rm max}\uparrow$     & 0.775 & 0.845 & 0.872 & 0.852  & 0.877 & 0.887 & 0.886 &0.896 &0.873 &0.903 &0.895 &0.893 &0.897 & 0.898 & \bf{0.914} \\
		& $E_{\phi}^{\rm max}\uparrow$    & 0.853 & 0.913 & 0.924 & 0.915  & 0.926 & 0.930 & 0.927 &0.935 &0.916 &0.944 &0.936 &0.937 &0.936  &0.935  &\bf{0.947}  \\
		& $\mathcal{M}\downarrow$ & 0.100 & 0.085 & 0.059 & 0.079  & 0.053 & 0.051 & 0.051 &0.043   &0.051 &0.043  &0.043 &0.047 &0.044 &0.045  &\bf{0.036} \\
		\midrule
		\multirow{4}{*}{\begin{sideways}\textit{NLPR}\end{sideways}}
		& $S_{\alpha}\uparrow$ & 0.799 & 0.860 & 0.874 & 0.856 &0.888 & 0.905 & 0.899 &0.914 &0.881 &0.925 &0.920 &0.908 &0.915 &0.915  &\bf{0.930} \\
		& $F_{\beta}^{\rm max}\uparrow$    & 0.771 & 0.825 & 0.841 & 0.815 &0.867 & 0.885 & 0.879 &0.896 &0.881 &0.916 &0.903 &0.887 &0.896  &0.903  &\bf{0.919} \\
		& $E_{\phi}^{\rm max}\uparrow$  & 0.879 & 0.929 & 0.925 & 0.913 &0.932 & 0.945 & 0.947 &0.953 &0.945 & 0.962 &0.956 &0.945 &0.949 &0.953 &\bf {0.965} \\
		& $\mathcal{M}\downarrow$ & 0.058 & 0.056 & 0.044 & 0.059 & 0.036 & 0.033 & 0.031 &0.026 &0.028 & 0.022 &0.025 &0.031 &0.027  &0.028 &\bf{0.022}  \\
		\midrule
		\multirow{4}{*}{\begin{sideways}\textit{STERE}\end{sideways}}
		& $S_{\alpha}\uparrow$& 0.825 & 0.848 & 0.875 & 0.873  & 0.879 & 0.889 & 0.886 &0.893  &0.879 &0.905 &0.903 &0.908 &0.895  &0.901  &\bf{0.911} \\
		& $F_{\beta}^{\rm max}\uparrow$   & 0.823 & 0.831 & 0.860 & 0.863  & 0.874 & 0.878 & 0.886 &0.889  &0.879 &0.901 &0.899 &0.905 &0.893 &0.892  &\bf{0.906} \\
		& $E_{\phi}^{\rm max}\uparrow$ & 0.887 & 0.912 & 0.925 & 0.927  & 0.925 & 0.929 & 0.938 &0.936 &0.928 &0.946 &0.944 &\bf{0.949} &0.939 &0.937  &0.947 \\
		& $\mathcal{M}\downarrow$   & 0.075 & 0.086 & 0.064 & 0.068  & 0.051 & 0.054 & 0.047 &0.044  &0.044 &0.042 &0.039 &0.040 &0.043  &0.043  &\bf{0.037} \\
		\midrule
		\multirow{4}{*}{\begin{sideways}\textit{RGBD135}\end{sideways}}
		& $S_{\alpha}\uparrow$ & 0.770 & 0.863 & 0.842 & 0.848  & 0.872 & 0.904 & 0.900 &0.904 &0.884 &0.929 &0.934 &0.909 &0.931 &0.924 &\bf{0.935}  \\
		& $F_{\beta}^{\rm max}\uparrow$  & 0.728 & 0.844 & 0.804 & 0.822  & 0.846 & 0.885 & 0.888 &0.884 &0.870 & 0.919 &\bf{0.930} &0.895 &0.922  &0.914  &0.929  \\
		& $E_{\phi}^{\rm max}\uparrow$ & 0.881 & 0.932 & 0.893 & 0.928  & 0.923 & 0.946 & 0.943 &0.941 &0.920 &0.968 &\bf{0.976} &0.945 &0.970  &0.966  &0.972 \\
		& $\mathcal{M}\downarrow$ & 0.068 & 0.055 & 0.049 & 0.065  & 0.038 & 0.030 & 0.030 &0.026 &0.029 & 0.022 &0.019 &0.028 &0.019 &0.023  &\bf{0.019} \\
		\midrule
		\multirow{4}{*}{\begin{sideways}\textit{DUTLF-D}\end{sideways}}
		& $S_{\alpha}\uparrow$ &0.468 &0.831 &0.801 &0.791 &0.749 &0.775 &0.889 &0.915 &0.885 &0.913 &0.863 &0.919 &0.912 &0.899 &\bf{0.932}\\
		& $F_{\beta}^{\rm max}\uparrow$ &0.357  &0.823 &0.771 &0.767 &0.718 &0.742 &0.898 &0.924 &0.892 &0.916 &0.857 &0.927 &0.914 & 0.906 &\bf{0.939}\\
		& $E_{\phi}^{\rm max}\uparrow$ &0.638 &0.899 &0.856 &0.859 &0.811 &0.834 &0.933 &0.951 &0.930 &0.949 &0.904 &0.956 &0.943 &0.940 &\bf{0.960}\\
		& $\mathcal{M}\downarrow$ &0.229 &0.097 &0.100  &0.113 &0.099 &0.097 &0.048 &0.033 &0.042 &0.039 &0.056 &0.033 &0.037 &0.043 &\bf{0.031}\\
		\midrule
		
		\multirow{4}{*}{\begin{sideways}\textit{SIP}\end{sideways}}
		& $S_{\alpha}\uparrow$ & 0.720 & 0.716 & 0.842 & 0.833 &0.850 & 0.864 & 0.806 &0.874 &0.826 & 0.879 &0.875 &0.858 &0.867 &0.875  &\bf{0.885} \\
		& $F_{\beta}^{\rm max}\uparrow$ & 0.712 & 0.694 & 0.838 & 0.818 &0.851 & 0.861 & 0.821 &0.880  &0.832 & 0.885 &0.879 &0.867 &0.871  &0.876  &\bf{0.889} \\
		& $E_{\phi}^{\rm max}\uparrow$ & 0.819 & 0.829 & 0.901 & 0.897 &0.903 & 0.910 & 0.875 &0.921 &0.890 &0.923 &0.919 &0.913 &0.907 &0.918  &\bf{0.924}\\
		& $\mathcal{M}\downarrow$  & 0.118 & 0.139 & 0.071 & 0.086 &0.064 & 0.063 & 0.085 &0.053 &0.070 &0.051 &0.051 &0.063 &0.060 &0.054  &\bf{0.048}\\
		\bottomrule
		\hline
	\end{tabular}
	\vspace{-8pt}
\end{table*}

\noindent \textbf{Channel-Modality Attention Module.} For generally enhancing features in a 3D decoder, we propose the Channel-Modality Attention module (CMA), which is inspired by  the squeeze-excitation attention block~\cite{hu2018squeeze}. The underlying purpose is to learn different attention weights \emph{considering both channel and temporal dimensions}. As shown in Fig. \ref{label2}, suppose the input is a 4D tensor with dimension $T\times H\times W\times C$. Firstly, the tensor is reshaped to $H\times W \times (C*T)$ to combine the modality information into the channel dimension. Next, the typical channel attention mechanism \cite{hu2018squeeze} is applied to the reshaped features as shown in Fig. \ref{label2}, and finally the attended feature tensor is reshaped back from $(H\times W \times (C*T))$ to $T\times H\times W\times C$ and then added with the original tensor to form a residual attention manner. Our experimental results show that CMA outperforms the naive 3D squeeze-excitation block and is more suitable for our framework.

\section{Experiments and Results}

\begin{figure*}
	\centering
	\includegraphics[width=1.0\linewidth]{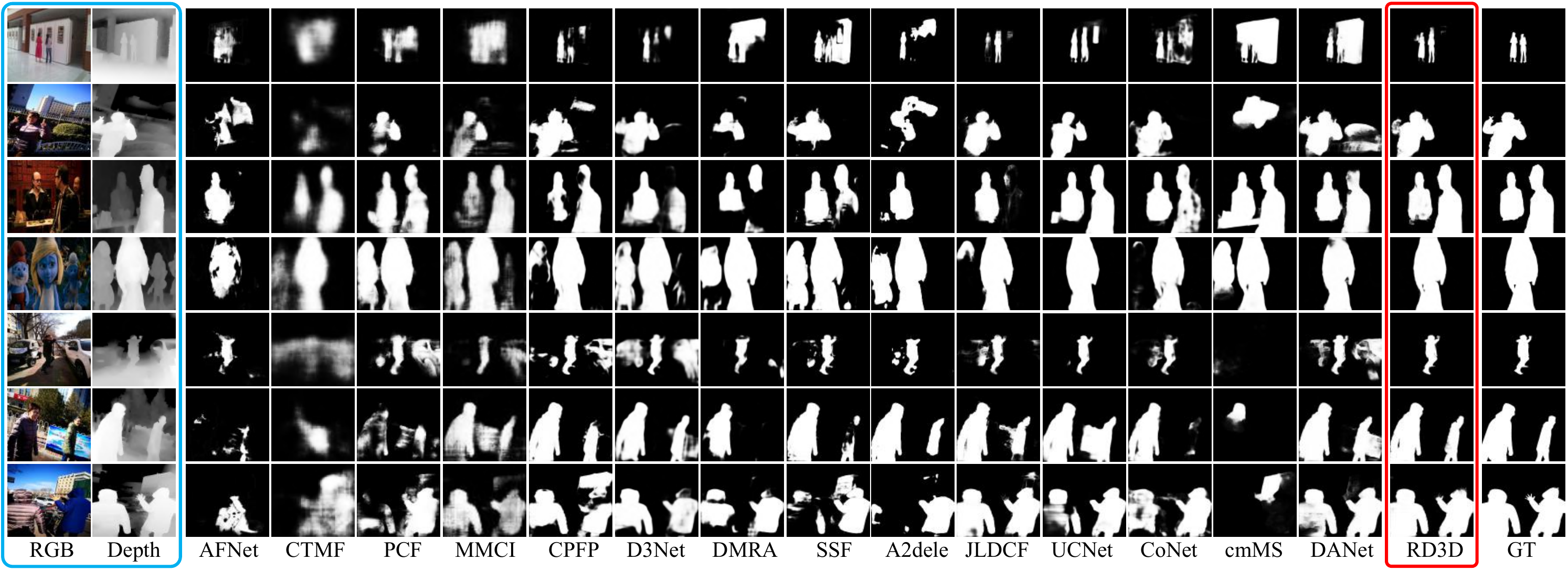}
	\vspace{-0.6cm}
	\caption{\small Qualitative comparisons of \ourmodel~with state-of-the-art (SOTA) methods. ``GT'' indicates the ground truth.}\vspace{-0.3cm}
	\label{fig4}
\end{figure*}

\subsection{Datasets and Metrics}
We evaluate our \ourmodel~on six popular public datasets having paired RGB and depth images, including: NJU2K (1,985 pairs) \cite{ju2014depth}), NLPR (1,000 pairs) \cite{peng2014rgbd}, STERE (1,000 pairs) \cite{niu2012leveraging}, DES (135 pairs, also called the RGBD135 dataset in some previous works) \cite{cheng2014depth}, SIP (929 pairs) \cite{fan2020rethinking} and DUTLF-D (1,200 pairs) \cite{piao2019depth}. Following \cite{chen2018progressively,chen2019multi,han2017cnns}, we use the same 1,485 pairs from NJU2K and 700 pairs from NLPR for training. The remaining pairs are used for testing. Specially, on the latest DUTLF-D dataset, we follow the same protocols as in \cite{piao2019depth,zhao2020single,piao2020a2dele,li2020rgb,ji2020accurate} to add additional 800 pairs from DUTLF-D for training and test on the remaining 400 pairs. In summary, our training set contains 2,185 paired RGB and depth images except when testing is conducted on DUTLF-D.

We use the newly proposed S-measure ($S_\alpha$) \cite{fan2017structure} and E-measure ($E_\phi$) \cite{fan2018enhanced}, as well as the generally agreed F-measure ($F_\beta$) \cite{borji2015salient} and Mean Absolute Error ($\mathcal{M}$) \cite{perazzi2012saliency} as evaluation metrics for comparing performance of different models. These four metrics provide comprehensive and reliable evaluation results and have been adopted by many previous works. Following \cite{fu2020jl}, we report the maximum F-measure ($F_\beta^{\rm max}$) and maximum E-measure ($E_\phi^{\rm max}$) scores.

\subsection{Implementation Details}
\subsubsection{3D ResNet}
We implement our 3D ResNet encoder based on the 2D ResNet \cite{ResNET}. We replace all 2D kernels in the ResNet-50 with their 3D versions and the 3D kernel weights are initialized by the 2D weights pre-trained on ImageNet \cite{russakovsky2015imagenet} in a centralized initialization manner \cite{girdhar2018detect}. We reduce the channel numbers of different side outputs to a fixed number $32$ in the channel reduction (CR) modules.

\begin{table*}[t]
  \centering
  \footnotesize
  \caption{\small Comparisons of different backbone strategies on four large datasets. The results of our \ourmodel~are highlighted in \textbf{bold}. Detailed analysis can be found in Section ``Ablation Studies: Backbone Strategies''.}\vspace{-0.5cm}
  \vspace{8pt}
  \scriptsize
  \renewcommand{\arraystretch}{0.87}
  \renewcommand{\tabcolsep}{0.35mm}
  \begin{tabular}{c|c|c|cccc|cccc|cccc|cccc}
  \Xhline{0.8pt}
  \multirow{2}{*}{Architecture}& Speed &Size  & \multicolumn{4}{c|}{NLPR (500 pairs)} & \multicolumn{4}{c|}{NJU2K (300 pairs)} & \multicolumn{4}{c|}{STERE (1000 pairs)}  & \multicolumn{4}{c}{SIP (929 pairs)}\\
  \cline{4-19}
   &(fps)&(MB) &\itshape{S}$_{\alpha}\uparrow$  & $\mathcal{M}\downarrow$ & $F_{\beta}^{\rm max}\uparrow$ &$E_{\phi}^{\rm max}\uparrow$ & \itshape{S}$_{\alpha}\uparrow$ & $\mathcal{M}\downarrow$ & $F_{\beta}^{\rm max}\uparrow$ &$E_{\phi}^{\rm max}\uparrow$ & \itshape{S}$_{\alpha}\uparrow$ & $\mathcal{M}\downarrow$ & $F_{\beta}^{\rm max}\uparrow$ &$E_{\phi}^{\rm max}\uparrow$&\itshape{S}$_{\alpha}\uparrow$ & $\mathcal{M}\downarrow$ & $F_{\beta}^{\rm max}\uparrow$ &$E_{\phi}^{\rm max}\uparrow$\\
  \Xhline{0.8pt}
   Input Fusion &47.4 &94.4 &.919 &.027 &.901 &.953 &.904 &.043 &.904 &.937 &.892 &.047 &.885 &.935 & .876 &.053 &.879 &.917 \\
   Two-stream &32.5 &200.0 &.929 &.023 &.918 &.962 &.913 &.039 &.911 &.944 & 907 &.040 &.899 &.941  &.878 &.052 &.881 &.923\\
   Siamese  &46.4 &94.4 & .927 &.024 &.917 &.959 &.915 &.037 &.913 &.946 &.904 &.041 &.898 &.939 &.867 &.057 &.867 &.905 \\
   \textbf{RD3D} &\textbf{45.6} & \textbf{180.8} &\textbf{.930} &\textbf{.022} &\textbf{.919} &\textbf{.965} &\textbf{.916} &\textbf{.036} &\textbf{.914} &\textbf{.947} &\textbf{.911}  &\textbf{.037} &\textbf{.906} &\textbf{.947} &\textbf{.885} &\textbf{.048} &\textbf{.889} &\textbf{.924} \\
  \Xhline{0.8pt}
  \end{tabular}
  \label{Tab4:ablation-withother}
  \vspace{-5pt}
\end{table*}

\subsubsection{Training and Testing Settings}
Our framework is implemented based on PyTorch~\cite{paszke2019pytorch} on a workstation with 4 NVIDIA 1080Ti GPUs. During training, we adopt the Adam optimizer with an initial learning rate of 0.0001, which is decayed by a cosine learning rate scheduler. The weight decay is set to 0.001. The data is first resized to $[352, 352]$ and then augmented by random horizontal flip and multi-scale transformation with the scale of \{256, 352, 416\}. We train for 100 epochs on 4 GPUs with the batch size equals to 10 per GPU, and the total training time is about 6 hours. The model after the last epoch is used for inference. Regarding the supervision, we calculate the typical binary cross-entropy loss. During testing, an image of arbitrary size is first resized to $[352, 352]$ and the predicted saliency map is resized back to its original size.
\subsection{Comparisons with SOTAs}
We compare \ourmodel~with 14 SOTA deep RGB-D SOD models, including AFNet~\cite{wang2019adaptive}, CTMF~\cite{han2017cnns}, PCF~\cite{chen2018progressively}, MMCI~\cite{chen2019multi}, CPFP~\cite{zhao2019contrast}, D3Net~\cite{fan2020rethinking}, DMRA~\cite{piao2019depth}, SSF~\cite{zhang2020select}, A2dele~\cite{piao2020a2dele}, JL-DCF~\cite{fu2020jl}, UCNet \cite{zhang2020uc,zhang2020uncertainty}, CoNet~\cite{ji2020accurate}, cmMS~\cite{li2020rgb} and DANet~\cite{zhao2020single}. Quantitative results are shown in Table~\ref{table:QuantitativeResults}. It can be seen that compared with other methods, our results have notable improvement on the six datasets, advancing the best scores obtained by SOTA models by an average of 0.68\%/0.50\% on $S_{\alpha}$/$F_\beta^{\rm max}$. We show visualization results of \ourmodel~and other methods in Fig. \ref{fig4}. In the global view, the detection of \ourmodel~is more accurate. In the detailed view, \emph{e.g.}, in the first row of Fig. \ref{fig4}, only \ourmodel~can accurately identify the two people as the foreground. In general, the decent qualitative performance of \ourmodel~is consistent with the quantitative analysis.
\begin{table*}[t]
  \centering
  \footnotesize
  \caption{\small Ablation results on four large datasets. The results of our \ourmodel~are in \textbf{bold}. Details about Model-1$\sim$Model-4 and detailed analysis can be found in Section ``Ablation Studies: Other Modules''.}\vspace{-0.5cm}
  \vspace{5pt}
  \scriptsize
  \renewcommand{\arraystretch}{0.9}
  \renewcommand{\tabcolsep}{0.55mm}
  \begin{tabular}{c|c|c|cccc|cccc|cccc|cccc}
  \Xhline{0.8pt}
  \multirow{2}{*}{Model} &Speed &Size & \multicolumn{4}{c|}{NLPR (500 pairs)} & \multicolumn{4}{c|}{NJU2K (300 pairs)} & \multicolumn{4}{c|}{STERE (1000 pairs)} & \multicolumn{4}{c}{SIP (929 pairs)}\\
  \cline{4-19}
  &(fps)&(MB)&\itshape{S}$_{\alpha}\uparrow$  & $\mathcal{M}\downarrow$ & $F_{\beta}^{\rm max}\uparrow$ &$E_{\phi}^{\rm max}\uparrow$ & \itshape{S}$_{\alpha}\uparrow$ & $\mathcal{M}\downarrow$ & $F_{\beta}^{\rm max}\uparrow$ &$E_{\phi}^{\rm max}\uparrow$ & \itshape{S}$_{\alpha}\uparrow$ & $\mathcal{M}\downarrow$ &$F_{\beta}^{\rm max}\uparrow$ &$E_{\phi}^{\rm max}\uparrow$ &\itshape{S}$_{\alpha}\uparrow$ & $\mathcal{M}\downarrow$ & $F_{\beta}^{\rm max}\uparrow$ &$E_{\phi}^{\rm max}\uparrow$\\
  \Xhline{0.8pt}
 DANet &32.0 &106.7 &.915 & .028 &.903 &.953 &.899 &.045 &.898 &.935 &.901 &.043 &.892 &.937 &.875 &.054 &.876 &.918\\
 JL-DCF &9.0 &520.0 &.925 &.022 &.916 &.962 &.903 &.043 &.903 &.944 &.905 &.042 &.901 &.946 &.879 &.051 &.885 &.923\\

   \textbf{RD3D} &\textbf{45.6} &\textbf{180.8} &\textbf{.930} &\textbf{.022} &\textbf{.919} &\textbf{.965} &\textbf{.916} &\textbf{.036} &\textbf{.914} &\textbf{.947} &\textbf{.911}  &\textbf{.037} &\textbf{.906} &\textbf{.947} &\textbf{.885} &\textbf{.048} &\textbf{.889} &\textbf{.924} \\
Model-1  &52.5 &180.5 &.913 &.031 &.894 &.949 &.906 &.043 &.898 &.940 &.897 &.049 &.884 &.935 &.873 &.059 &.867 &.915\\
Model-2  &50.2 &180.5 &.918 &.028 &.899 &.949 &.913 &.040 &.913 &.944 &.906 &.042 &.897 &.940  &.878 &.053 &.882 &.919\\
Model-3  &45.8 &180.7 &.921 &.027 &.904 &.949  &.914 &.039 &.913 &.942 &.907  &.042 &.897 &.939  &.866 &.059 &.864 &.901 \\
Model-4  &40.4 &219.1 &.931 &.022 &.921 &.965    &.920 &.034 &.923 &.952  &.908 &.039 &.901 &.944  &.883  &.048  &.890  &.924\\
   \Xhline{0.8pt}
  \end{tabular}
  \label{Tab4:ablation-method}
\end{table*}

\vspace{-0.2cm}
\subsection{Ablation Studies: Backbone Strategies}
\begin{figure*}
	\centering
	\includegraphics[width=0.87\linewidth]{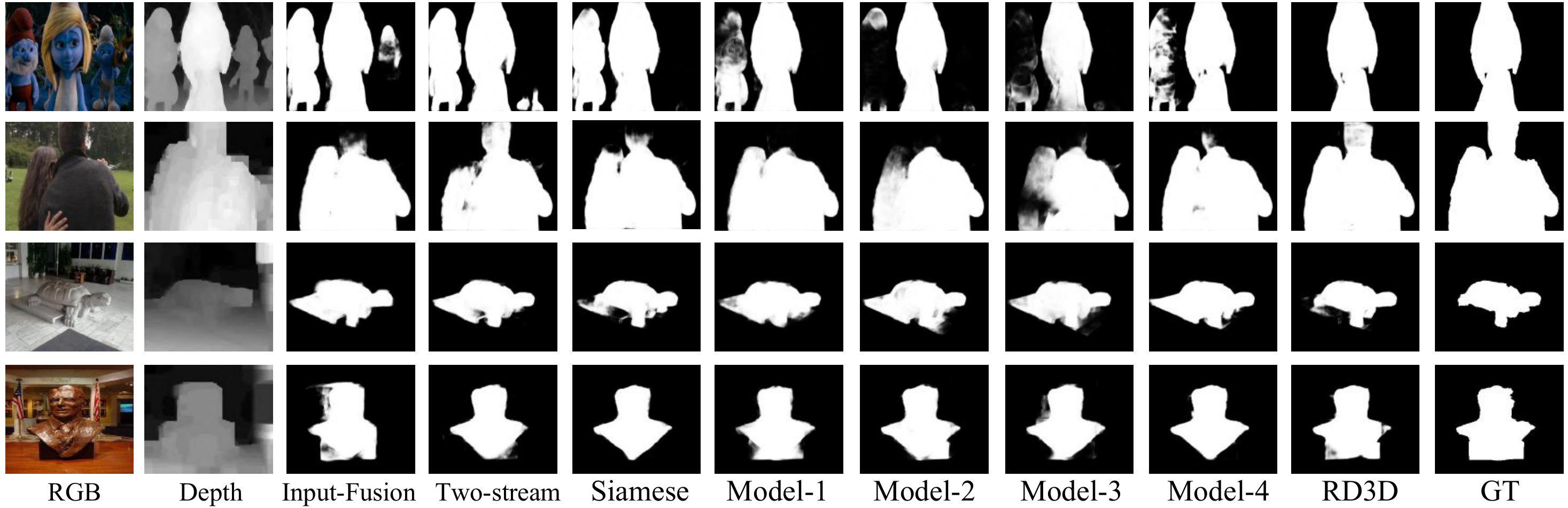}
	\vspace{-0.35cm}
	\caption{\small Visual results for other architectures and ablation studies. In general, \ourmodel~provides the closest results to the GT.}\vspace{-0.3cm}
	\label{fig5}
\end{figure*}
To verify the effectiveness of pre-fusion in the backbone via 3D CNNs, we compare the four backbone strategies shown in Fig. \ref{related} (a)-(d). Our method belongs to Fig. \ref{related} (d), and we implement Input Fusion Network (Fig. \ref{related} (c)), Two-stream Network (Fig. \ref{related} (a)), and Siamese Network (Fig. \ref{related} (d)) by switching the encoder part of \ourmodel. Note that the main difference lies in the way the encoder deals with multi-modal inputs. For fair comparison, we keep the decoder the same.
We implement Input Fusion Network by first concatenating the RGB and depth images in the channel dimension and then fusing them by the first convolution layer in the 2D ResNet. Since the input shape is inconsistent with the original ResNet, we modify the first convolution layer and later repeat the encoder outputs in the temporal axis to enforce input $T=2$ for the decoder. For Two-stream Network, we use two 2D ResNets to extract hierarchical features separately. Likewise, features are then concatenated in the temporal axis. The Siamese Network is implemented by a shared 2D ResNet for RGB and depth, while keeping other settings the same.

Table~\ref{Tab4:ablation-withother} shows experimental results on four large datasets including NLPR, NJU2K, STERE and SIP.
As can be seen, \ourmodel~based on 3D CNNs outperforms the other three strategies by a notable margin. The Input Fusion Network performs worst though its model size is small, because multi-modality inputs are fused too naively, leading to insufficient extraction of multi-modal information. Besides, the Two-stream Network and Siamese Network are comparable to each other, but both are worse than our strategy. This clearly demonstrates the effectiveness of pre-fusion in the backbone through 3D convolutions. Regarding the model speed and size of our scheme, they are almost equal to those of the Two-stream Network, but our numbers are slightly better. Some visual examples are shown in Fig. \ref{fig5}.

\subsection{Ablation Studies: Other Modules}
We take the full model of \ourmodel~as the reference and conduct thorough ablation studies by replacing or removing the key components. The full version is denoted as RD3D (3D ResNet+CMA +RBPP), where ``CMA'' and ``RBPP'' refer to the usage of CMA modules and RBPP. Firstly, we construct a baseline ``Model-1'' (3D ResNet) by removing CMA modules and RBPP. Thus, the decoder of this model is just a plain 3D UNet decoder.
Secondly, to validate the effectiveness of the rich back-projection paths (RBPP), we realize ``Model-2'' (3D ResNet+CMA) by removing all the back-projection paths.
Thirdly, to demonstrate the benefit of channel-modality attention modules (CMA), we implement ``Model-3'' (3D ResNet+CA+RBPP), which replaces all CMA modules with naive squeeze-excitation channel attention modules \cite{hu2018squeeze}, namely only channel attention is considered and during the squeeze operation, the global pooling is applied to the other three dimensions. 
Lastly, to investigate the proposed CMA modules, we also construct ``Model-4'' (3D ResNet+CMA*+RBPP), where ``CMA*'' means moving CMA from the decoder to the encoder stage. CMA modules are inserted into the ResNet backbone in a way as suggested by \cite{hu2018squeeze}. Results of the above ablation studies are reported in Table~\ref{Tab4:ablation-method}, where two SOTA models DANet and JL-DCF are listed also. Visual comparisons are presented in Fig. \ref{fig5}. The following observations can be achieved.

\noindent \textbf{Effectiveness of the Baseline Model.} Without bells and whistles, the baseline model ``Model-1'' performs favorably against the two latest SOTA models DANet and JL-DCF, showing the potentials of using 3D convolutions for achieving effective cross-modality feature aggregation. Note that this baseline model consists of only basic 3D convolutions without any other augmentation.

\noindent \textbf{Effectiveness of RBPP.} Comparing between ``Model-2'' and the full \ourmodel~in Table \ref{Tab4:ablation-method} shows that removing the RBPP leads to consistent performance degeneration. This implies that taking use of all information from higher-resolution levels is beneficial, especially to our framework where the back-projection paths contain rich multi-level modality-aware information.

\noindent \textbf{Effectiveness of CMA.} Comparing ``Model-3'' to \ourmodel~in Table~\ref{Tab4:ablation-method}, one see that when the CMA modules are replaced, the performance drops, demonstrating that our channel-modality attention mechanism can enhance the final prediction and is probably more suitable for our fully 3D CNNs-based framework. Comparing ``Model-2'' to ``Model-1'', without RBPP, the improvement from adding CMA is still notable. This implies that combining RBPP and CMA is a reasonable and effective design, which results in substantial enhancement. 
In addition, ``Model-4'' achieves slightly better benchmark results than \ourmodel, showing that the proposed CMA modules can also work on the backbone. However, moving CMA to the encoder leads to slightly the higher computation and model size as in Table~\ref{Tab4:ablation-method}, because much more CMA modules have been deployed. Since using attention modules in the backbone is usually not adopted by the previous works, for fair comparison, we opt to deploy CMA in the decoder of \ourmodel.

\section{Conclusion} 
We propose a novel RGB-D SOD framework called \ourmodel, which is based on 3D CNNs and conducts cross-modal feature fusion in a progressive manner. \ourmodel~first utilizes 3D convolutions for pre-fusion between RGB and depth, and then conduct explicit fusion of modality-aware features by a 3D decoder augmented with rich back-projection paths and channel-modality attention modules. Extensive experiments on six benchmark datasets demonstrate that \ourmodel, which is the first fully 3D CNNs-based RGB-D SOD model, performs favorably against existing SOTA approaches. Detailed ablation studies and discussions validate the key components of \ourmodel. In the future, we hope \ourmodel~could encourage more RGB-D SOD designs based on 3D CNNs.   

\vspace{-2pt}
\small{\vspace{.1in}\noindent\textbf{Acknowledgments.}\quad
This work was supported by the NSFC, under No. 61703077, 61773270, 61971005, the
Fundamental Research Funds for the Central Universities No. YJ201755.}

\bibliography{bibfile}
\end{document}